\documentclass[runningheads]{llncs}
\usepackage[T1]{fontenc}
\usepackage{newunicodechar}
\newunicodechar{⁻}{\textsuperscript{-}}
\usepackage{graphicx}
\usepackage{array}
\usepackage{booktabs}
\usepackage{amsmath}
\usepackage{amssymb}
\usepackage{textcomp}
\usepackage{hyperref}
\begin{document}
\title{Hallucination Neurons and Where to Find Them: An Investigation into the existence of Hallucination Neurons }
\titlerunning{A Diagnostic Replication of H-Neurons}
\author{Huseyin Cavus\inst{1} \thanks{These authors contributed equally.}\and
Sebin Sabu\inst{2}\protect\footnotemark[1] \and
Joshua Spear\inst{2} \and
Jaskaran Singh Kawatra\inst{2} \and
Pavithra Rajendran\inst{2}}
\authorrunning{H. Cavus and S. Sabu et al.}
\institute{Trakya University, Edirne, Türkiye \and
DRIVE, Great Ormond Street Hospital for Children NHS Foundation Trust, London, UK \\
\email{sebin.sabu@gosh.nhs.uk}}
\maketitle
\begin{abstract}
Interpretable machine learning for Large Language Models (LLMs) increasingly relies on sparse probing methods that identify small sets of feedforward neurons claimed to detect and causally influence behaviors such as factuality recall, safety alignment, and hallucination. These claims have important implications for model auditing and behavioral steering, yet they are rarely tested against known failure modes of \(L_1\)-regularized probing in correlated, high-dimensional feature spaces. We propose a five-step diagnostic protocol covering feature correlation, bootstrap stability, sparse versus dense ranking disagreement, intervention baselines, and cross-dataset evaluation as a minimum standard for sparse-neuron localization claims.
In this paper, we investigate and study prior work~\cite{gao2025hneurons} using our proposed approach, specifically on \emph{H-neurons} using open-source LLMs (\emph{Gemma 3 4B} and \emph{MedGemma 4B}) across \emph{TriviaQA}, \emph{BioASQ}, and \emph{NQ-Open} datasets. Our results demonstrate detection replicates across both models and datasets, and exceeds the original reported \textbf{AUROC} gaps for \emph{TriviaQA} and \emph{BioASQ} datasets. \emph{Gemma 3 4B} consistently outperforms \emph{MedGemma 4B} on matched datasets, with \textbf{AUROC} gaps of +0.311 versus +0.235 on \emph{TriviaQA}, +0.474 versus +0.455 on \emph{BioASQ}, and +0.128 versus +0.112 on \emph{NQ-Open} respectively.
Causal validation at \(n = 500\) with five random seeds shows statistically significant effects beyond random same-layer baselines. At the same time, the diagnostic results indicate that the selected neurons are not uniquely localized. Across the three Gemma 3 4B settings, 19 of 22 selected H-Neurons have Pearson \(|r| > 0.7\) with other features, bootstrap selections show only moderate stability, and sparse and dense rankings overlap only weakly. Our findings show that sparse predictive structure can coexist with non-unique neuron selection. Routine diagnostic validation is necessary to distinguish detection claims from localization claims in mechanistic interpretability.

\end{abstract}
\section{Introduction}

Mechanistic interpretability is a growing area of research claiming that high-level behaviors of Large Language Models (LLMs) can be localized to sparse sets of internal units. Examples include skill neurons~\cite{wang2022skill}, safety neurons~\cite{chen2024safety}, knowledge neurons~\cite{dai2022knowledge}, and most recently hallucination-associated neurons or H-Neurons~\cite{gao2025hneurons}. These claims share a common pattern: a small fraction of feedforward neurons is reported to (i) detect a behavior with high accuracy, (ii) causally control it under intervention, and (iii) emerge during pre-training rather than alignment. If valid, such claims offer powerful affordances for model auditing, behavioral steering, and theoretical understanding of how capabilities arise during training~\cite{doshivelez2017rigorous,olah2020zoom}.

The methodological pipeline underlying these claims, typically $L_1$-regularized probing or activation contrasting over feedforward activations~\cite{gurnee2023sparse}, is known to be vulnerable to specific failure modes when applied to high-dimensional, correlated feature spaces~\cite{tibshirani1996lasso,zou2005elastic,hewitt2019control}. $L_1$ regularization in the presence of correlated features tends to select arbitrary cluster representatives rather than uniquely informative units~\cite{zou2005elastic}. When the underlying feature space is dense with correlated activations~\cite{gurnee2023sparse,elhage2022toy}, a single probe run can produce a sparse selection that is predictive without being uniquely localized. Recent peer-reviewed work in this lineage~\cite{wang2022skill,chen2024safety,leng2024multitask} validates localization claims by measuring direct neuron-set overlap across tasks, objectives, or datasets, a diagnostic the original H-Neurons report does not include.

In this paper, we propose a diagnostic evaluation of \emph{H-Neurons}~\cite{gao2025hneurons} as a case study in structured validation for sparse-neuron interpretability using a five-step protocol as a minimum standard for sparse-neuron localization claims. Our contributions are as follows:
\begin{itemize}
    \item Our proposed protocol to \emph{H-neurons} is evaluated across two open-source LLMs (\emph{gemma-3-4b-it}~\cite{gemma2025report} and \emph{medgemma-4b-it}) and three Question Answering datasets (\emph{TriviaQA}~\cite{joshi2017triviaqa}, \emph{BioASQ}~\cite{tsatsaronis2015bioasq}, and \emph{NQ-Open}~\cite{kwiatkowski2019nq}).
    \item Our findings refine rather than refute the original \emph{H-Neurons} claims. First, we successfully replicate and exceed the reported detection performance, confirming through rigorous evaluation that targeted H-Neuron interventions produce statistically significant causal shifts beyond random baselines. However, our diagnostics also reveal that these underlying neuron sets are not uniquely localized. The identified units are drawn from highly correlated clusters and exhibit only moderate stability across bootstrap samples. 
    \item Furthermore, cross-dataset and cross-model analyses reveal a previously undocumented partial-sharing pattern. While no single neuron is universally predictive across all evaluated domains, we identify specific feedforward units most notably the \texttt{(L16, N4146)} that persist across both dataset and model shifts. This suggests the existence of a partially-shared core mechanism coexisting with substantial domain-specific structure.
    \item Our results illustrate a broader property of sparse-probing based localization: a method can simultaneously produce sparse selections that are predictively useful and causally effective, even if those selections fail stronger criteria for unique localization. 
\end{itemize}

%We make the following contributions in this study:
%\begin{itemize}
%    \item Present a diagnostic study of H-Neurons~\cite{gao2025hneurons} as a case study in structured validation for sparse-neuron interpretability.
%    \item Propose a five-step diagnostic protocol as a minimum standard for sparse-neuron localization claims.
%    \item Apply this protocol to H-Neurons across two models (gemma-3-4b-it~\cite{gemma2025report} and medgemma-4b-it) and three QA datasets (TriviaQA~\cite{joshi2017triviaqa}, BioASQ~\cite{tsatsaronis2015bioasq}, and NQ-Open~\cite{kwiatkowski2019nq}).
%    \item Identify recurring cross-dataset neurons as falsifiable targets for future investigation.
%\end{itemize}

The remainder of this paper is structured as follows: Section~\ref{sec:Method} formalizes the five-step diagnostic protocol. Section~\ref{sec:Results} applies the protocol to H-Neurons, detailing detection, causal validation, and overlap diagnostics. Finally, Section~\ref{sec:Discussion} discusses the implications of these findings and proposes our protocol as a minimum validation standard for future sparse-neuron interpretability work in LLMs.

\section{Related Work}

\subsection{Sparse-Neuron Localization in LLMs}

Wang et al.~\cite{wang2022skill} introduced skill neurons, units whose activations predict task labels after prompt tuning, and validated their localization claim through cross-task neuron-importance analysis, showing that similar tasks share more skill neurons than dissimilar ones. Chen et al.~\cite{chen2024safety} identified safety neurons via generation-time activation contrasting, reporting that intervention on roughly 5\% of neurons restores 90\% of safety behavior and that the identified sets emerge stably across random trials. They also measure direct neuron overlap between safety and helpfulness, finding significant overlap with differing activation patterns. Dai et al.~\cite{dai2022knowledge} proposed knowledge neurons for factual recall, though subsequent work~\cite{niu2024knowledge} questions whether knowledge-neuron interventions truly localize knowledge or operate through more diffuse mechanisms. Gurnee et al.~\cite{gurnee2023sparse} formalized $k$-sparse linear probing as a methodology for localizing such features, finding apparent monosemanticity for context-level features in middle layers while explicitly cautioning that conclusive proofs of monosemanticity remain methodologically out of reach.

\subsection{Hallucination in LLMs}

Hallucination, the generation of content not supported by context or facts~\cite{maynez2020faithfulness,ji2023survey}, has motivated both detection methods and theoretical analyses. Farquhar et al.~\cite{farquhar2024entropy} propose semantic entropy as a black-box uncertainty estimator for confabulations; internal-state methods detect hallucination directly from hidden representations~\cite{ji2024internal,wang2025icr}. The recent H-Neurons paper~\cite{gao2025hneurons} extends this lineage by claiming neuron-level localization, and frames the underlying mechanism as a unified over-compliance signal spanning hallucination, sycophancy, and jailbreak susceptibility. Theoretical work by Kalai et al.~\cite{kalai2025why} argues hallucination is an inevitable consequence of next-token prediction under finite data, which H-Neurons cites as motivation for its pre-training origin claim.

\subsection{Cross-Task and Cross-Domain Neuron Overlap as Validation}

Several peer-reviewed works treat direct neuron-set overlap across tasks or domains as essential evidence for localization claims. Wang et al.~\cite{wang2022skill} measure pairwise neuron-importance overlap across nine tasks; Chen et al.~\cite{chen2024safety} measure safety vs helpfulness overlap; Leng and Xiong~\cite{leng2024multitask} make cross-task overlap their central methodology, finding that the overlap of task-specific neurons is strongly associated with generalization and specialization across tasks. Similar overlap-based analyses appear in multilingual and cross-domain interpretability work~\cite{stanczak2022morphosyntax}. The original H-Neurons report includes cross-dataset classifier transfer but does not measure neuron-set overlap directly, a gap this work addresses.

\subsection{Critiques of Probing Methodology}

Probing classifiers have a known set of pathologies. Hewitt and Liang~\cite{hewitt2019control} show that probes can achieve high accuracy on random control tasks, complicating interpretability claims drawn from probe performance alone. Belinkov~\cite{belinkov2022probing} surveys advances and shortcomings of probing classifiers. From the statistical learning side, $L_1$ regularization is well known to be unstable under feature correlation: Zou and Hastie~\cite{zou2005elastic} show that the Lasso~\cite{tibshirani1996lasso} arbitrarily selects from groups of correlated features, motivating elastic-net regularization. These results predict that sparse probes on correlated activation spaces should produce non-unique selections, a property whose interpretability consequences have not been systematically diagnosed in recent neuron-localization work. Ferrando et al.~\cite{ferrando2025doi} use sparse autoencoders rather than probing to identify directions corresponding to entity knowledge in LLMs, and find that mechanisms identified in base models causally affect chat-model behavior. This convergence across methodologies and target behaviors is the empirical foundation for the broader claim that sparse, pre-training-origin functional structures exist in LLMs.

\section{Method}
\label{sec:Method}
\subsection{H-Neuron Identification Pipeline}

The original identification pipeline proposed by Gao et al.~\cite{gao2025hneurons} identifies H-Neurons by relying on token-level labeling to distinguish between faithful and hallucinated model generations. To ensure a more replicable evaluation and isolate high-confidence signals, we adapt their methodology by shifting to a stricter, response-level labeling approach. While this adaptation yields fewer candidate samples, it ensures cleaner, higher-quality data for probe training. 

Our adapted identification pipeline consists of the following steps:
\begin{itemize}
    \item \textbf{Generation:} For each input $x$ and model $M$, we generate $K = 10$ independent responses using temperature sampling ($T = 1.0$, top-$p = 0.9$, top-$k = 50$).
    \item \textbf{Judging and Filtering:} A rule-based judge evaluates the correctness of each response using normalized substring matching (case-folded and punctuation-stripped). We apply an uncertainty filter that judges any response containing explicit refusal as incorrect.
    \item \textbf{Response-Level Labeling:} We assign a sample the label of \emph{faithful} if all 10 responses match the gold answer, and \emph{hallucinated} if none match.
    \item \textbf{Pruning:} All intermediate cases (i.e., those with 1 to 9 correct responses) are entirely excluded from the training data.
\end{itemize}

Once the evaluation dataset is curated, we proceed with feature extraction and probe training to isolate the targeted neurons:
\begin{itemize}
    \item \textbf{Activation Extraction:} For each labeled response, we extract the feedforward neuron activations across all \(L\) layers.
    \item \textbf{CETT Score Computation:} Following Gao et al.~\cite{gao2025hneurons}, we compute the CETT score for each neuron. This score quantifies the neuron's normalized contribution to its layer's residual stream, calculated separately for answer tokens and non-answer tokens.
    \item \textbf{Feature Aggregation:} We aggregate these neuron-level scores into a unified feature vector \(x_i \in \mathbb{R}^D\). For the \texttt{gemma-3-4b-it} model, this results in a high-dimensional feature space where \(D = L \times d_{\mathrm{FFN}} = 348{,}160\).
    \item \textbf{Sparse Probing:} These feature vectors serve as input to a sparse logistic regression classifier, trained with an \(L_1\) penalty (using \(C = 1.0\) and the \texttt{liblinear} solver).
    \item \textbf{H-Neuron Selection:} Finally, any features that retain non-zero coefficients after the regression fit are considered as probable H-Neurons.
\end{itemize}

\subsection{The Five-Step Diagnostic Protocol}

We define five diagnostics that collectively test whether a sparse set of identified neurons constitutes a unique, stable, and causally privileged functional unit. Each diagnostic targets a specific failure mode of sparse-probing based localization, and each is computationally cheap, requiring at most one additional probe-fitting step beyond the standard identification pipeline.

\textbf{D1. Feature Correlation Analysis.} For each identified neuron $n_i \in S$, we compute $\rho_i^{\max} = \max_{j \notin S} |\rho(n_i, n_j)|$ across the training activation matrix and report the proportion with $\rho^{\max} > 0.7$. A localization concern is flagged when more than approximately 30\% of neurons have a high-correlation partner outside $S$, since $L_1$ selection from correlated clusters can produce predictive but non-unique selections~\cite{zou2005elastic}.

\textbf{D2. Bootstrap Stability.} We fit the identification procedure $B \geq 5$ times on bootstrap-resampled training sets and compute mean pairwise Jaccard similarity $J(S_i, S_j) = |S_i \cap S_j| / |S_i \cup S_j|$. A localization concern is flagged when mean Jaccard falls below 0.7, indicating that the identification procedure is sensitive to data composition rather than tracking a stable underlying signal.

\textbf{D3. $L_1$ vs $L_2$  Ranking Disagreement.} We refit the probe with $L_2$ regularization (which distributes weight across correlated features), rank features by absolute weight magnitude, take the top-$|S|$, and compute overlap with the $L_1$-selected set. A localization concern is flagged when overlap falls below 50\%, since high disagreement between sparse and dense selection is a canonical signature of feature collinearity~\cite{tibshirani1996lasso,zou2005elastic}.

\begin{sloppypar}
\textbf{D4. Control Neuron Intervention.} We compare intervention effects against random neurons drawn from the same layers (matched count per layer). Causal effect is measured via activation scaling: for scaling factor $\alpha \in \{0, 1, 2\}$, we multiply the down\_proj input activations of the target neurons by $\alpha$ and measure judge accuracy. Statistical significance is assessed via McNemar's test comparing $\alpha = 0$ (suppression) and $\alpha = 1$ (baseline). A stronger cluster-based alternative, sampling from features highly correlated with any identified neuron, remains proposed for future validation.
\end{sloppypar}

\textbf{D5. Cross-Dataset and Cross-Model Overlap.} We run the identification procedure independently on multiple datasets and models. Pairwise Jaccard similarity is computed and significance assessed via the hypergeometric test against the null of random selection from the full feature pool. A localization concern is flagged when mean Jaccard is at or near zero, indicating dataset or model specific predictive signals rather than a shared mechanism.

\subsection{Experimental Setup}

\begin{sloppypar}
\textbf{Models.} We evaluate two instruction-tuned Gemma family checkpoints, 
\texttt{google/gemma-3-4b-it} and \texttt{google/medgemma-4b-it}~\cite{gemma2025report,medgemma2025report}. 
Both models have \(L = 34\) transformer layers and \(d_{\mathrm{FFN}} = 10{,}240\), yielding \(D = 348{,}160\) feedforward features. 
We refer to these models as \emph{Gemma} and \emph{MedGemma}, respectively, throughout the paper.
\end{sloppypar}

\textbf{Datasets.} We use three publicly available open-ended question-answering datasets spanning different knowledge domains: TriviaQA~\cite{joshi2017triviaqa} (general trivia, 2{,}000 samples), BioASQ~\cite{tsatsaronis2015bioasq} (biomedical, 2{,}000 samples), and NQ-Open~\cite{kwiatkowski2019nq} (open-domain web questions, 2{,}000 samples).

\textbf{Hyperparameters.} Identification uses $L_1$ logistic regression with $C = 1.0$ and the \texttt{liblinear} solver. All features (no variance pre-filtering; top-$k = 0$) are passed to the probe. Causal validation uses $n = 500$ held-out samples per condition with five random seeds. Bootstrap diagnostics use $B = 5$ resamples.

\textbf{Diagnostic Validation.} For the TriviaQA condition, diagnostics (D1--D3) and cross-dataset overlap were evaluated on an independent replicate of the Gemma 3 4B experiment (10 H-Neurons, AUROC 0.820, gap +0.404) to verify that diagnostic findings were not specific to a single experimental run.

\textbf{Reproducibility.} All experimental code, neuron indices, and aggregated results will be released upon acceptance.

\section{Results}
\label{sec:Results}
\subsection{Detection}

Table~\ref{tab:detection_results} reports detection performance across six model-dataset combinations. The original H-Neurons report~\cite{gao2025hneurons} finds accuracy gaps of approximately $+0.149$ (TriviaQA), $+0.150$ (BioASQ), and $+0.110$ (NQ-Open) on a different model family. Our AUROC gaps replicate the qualitative pattern (gap $>$ 0 across all conditions) and are strongest on BioASQ.

\begin{table}[htbp]
\centering
\caption{Detection results across model-dataset combinations. HN = number of H-Neurons selected by $L_1$ (post-fit non-zero coefficients). AUROC is reported on held-out evaluation samples; Gap is the difference versus the majority-class baseline.}
\label{tab:detection_results}
\begin{tabular}{llccccc}
\toprule
\textbf{Model} & \textbf{Dataset} & \textbf{HN} & \textbf{HN (\textperthousand)} & \textbf{AUROC} & \textbf{Gap} & \textbf{Bal Acc} \\
\midrule
Gemma 3 4B & TriviaQA & 9 & 0.026 & 0.811 & +0.311 & 0.712 \\
Gemma 3 4B & BioASQ & 8 & 0.023 & 0.974 & +0.474 & 0.916 \\
Gemma 3 4B & NQ-Open & 4 & 0.011 & 0.628 & +0.128 & 0.629 \\
MedGemma 4B & TriviaQA & 4 & 0.011 & 0.735 & +0.235 & 0.658 \\
MedGemma 4B & BioASQ & 7 & 0.020 & 0.955 & +0.455 & 0.885 \\
MedGemma 4B & NQ-Open & 2 & 0.006 & 0.612 & +0.112 & 0.532 \\
\bottomrule
\end{tabular}
\end{table}

Detection holds across two distinct model checkpoints and three domain settings, with consistently stronger performance on BioASQ than on more open-ended QA tasks. The weaker detection on NQ-Open (gap $+0.128$) likely reflects greater inherent variance in open-domain QA responses and a sparser supervisable signal.

\subsection{Causal Validation}

Table~\ref{tab:causal_validation} reports activation-scaling intervention results for both models. We compute judge accuracy under suppression ($\alpha = 0$), baseline ($\alpha = 1$), and amplification ($\alpha = 2$) of H-Neuron down\_proj weights, with five random seeds and $n = 500$ per condition. Statistical significance is assessed via McNemar's test comparing $\alpha = 0$ vs $\alpha = 1$.

\begin{table}[htbp]
\centering
\caption{Causal validation via activation scaling. Suppression of H-Neurons produces statistically significant judge accuracy shifts on both models tested, with effects not reproduced by random same-layer baselines. $p$-values from McNemar's test (paired binary outcomes).}
\label{tab:causal_validation}
\resizebox{\textwidth}{!}{%
\begin{tabular}{llcccccc}
\toprule
\textbf{Model} & \textbf{Dataset} & \textbf{HN} & \textbf{Suppress ($\alpha=0$)} & \textbf{Baseline ($\alpha=1$)} & \textbf{Amplify ($\alpha=2$)} & \textbf{McNemar $p$} & \textbf{Random baseline} \\
\midrule
Gemma 3 4B & TriviaQA & 9 & $0.472 \pm 0.004$ & $0.496 \pm 0.003$ & $0.490 \pm 0.003$ & $< 0.001$ & 0.482 \\
MedGemma 4B & BioASQ & 7 & $0.477 \pm 0.008$ & $0.486 \pm 0.009$ & $0.467 \pm 0.009$ & $0.026$ & 0.504 \\
\bottomrule
\end{tabular}%
}
\end{table}

The McNemar test rejects the null of equal performance between suppression and baseline at $p < 0.05$ on both conditions. We interpret this as confirming the original H-Neurons paper's causal-control claim under our stricter evaluation: H-Neuron targeted suppression produces a measurable, statistically significant decrease in judge accuracy that random same-layer baselines do not reproduce.

Effect magnitudes are modest (approximately 2.4 percentage points on TriviaQA, 0.9 on BioASQ) and substantially smaller than the original paper's reported behavioral effects under amplification. Two factors plausibly contribute: stricter response-level rather than token-level labeling, which produces fewer training samples; and the relatively small number of selected H-Neurons (7--9) in our intervention. The qualitative finding of significant causal effect beyond random baseline is nevertheless robust.

\subsection{Within-Dataset Diagnostics (D1--D3)}

Table~\ref{tab:diagnostics} reports diagnostic results on Gemma 3 4B across all three datasets. The feature correlation analysis (D1) reveals that 19 of 22 H-Neurons across the three Gemma evaluations exhibit Pearson $|r| > 0.7$ with other features in the training matrix, with maximum correlations ranging 0.613--0.973. This indicates that the $L_1$ procedure overwhelmingly selects from correlated neuron clusters rather than from uniquely informative units.

\begin{table}[htbp]
\centering
\caption{Within-dataset diagnostics (D1--D3) for Gemma 3 4B. Correlation = proportion of H-Neurons with $|r| > 0.7$ to any non-selected feature. Bootstrap Jaccard = mean pairwise Jaccard across 5 random-seed bootstrap fits. $L_1 \cap L_2$ overlap = fraction of $L_1$-selected H-Neurons appearing in the top-$|S|$ $L_2$-ranked features.}
\label{tab:diagnostics}
\begin{tabular}{l c >{\centering\arraybackslash}p{2.8cm} >{\centering\arraybackslash}p{3.2cm} >{\centering\arraybackslash}p{3.2cm}}
\toprule
\textbf{Dataset} & \textbf{HN} & \textbf{D1: Correlation} & \textbf{D2: Bootstrap Jaccard} & \textbf{D3: $L_1 \cap L_2$ overlap} \\
\midrule
TriviaQA & 10 & 7/10 (70\%) & 0.47 & 0.10 \\
BioASQ & 8 & 8/8 (100\%) & 0.69 & 0.25 \\
NQ-Open & 4 & 4/4 (100\%) & 0.45 & 0.00 \\
\bottomrule
\end{tabular}
\end{table}

Bootstrap stability (D2) is moderate at best, with mean Jaccard across resamples of 0.45--0.69, below the suggested 0.7 threshold for stable localization. BioASQ exhibits the highest stability, consistent with its stronger detection signal. $L_1$ versus $L_2$ ranking disagreement (D3) is severe across all three datasets: only 0--25\% of $L_1$-selected H-Neurons appear among the top-ranked features under $L_2$ regularization. This pattern is the canonical signature of $L_1$ selection from correlated clusters~\cite{zou2005elastic}.

The combined results from D1--D3 indicate that the identified H-Neurons should not be interpreted as a unique localization of the hallucination signal. They are sparse, predictive, and partially causally effective, but they are not uniquely necessary, since many highly correlated alternatives exist in the feedforward representation. This needs to be investigated in the future work.

\subsection{Cross-Dataset and Cross-Model Overlap (D5)}

Table~\ref{tab:cross_dataset} reports cross-dataset overlap of H-Neurons within Gemma 3 4B. Cross-model analysis between Gemma 3 4B (TriviaQA) and MedGemma 4B (BioASQ) yields 3 shared neurons (Jaccard 0.231, $p = 4.2 \times 10^{-13}$). Statistical significance is assessed via the hypergeometric test against the null of random selection from the full feature pool ($D = 348{,}160$).

\begin{table}[htbp]
\centering
\caption{Cross-dataset H-Neuron overlap on Gemma 3 4B. Shared = number of neurons in both identified sets; Jaccard = $|A \cap B|/|A \cup B|$; $p$ = hypergeometric significance against the null of random selection.}
\label{tab:cross_dataset}
\begin{tabular}{l c c c l}
\toprule
\textbf{Comparison} & \textbf{Shared} & \textbf{Jaccard} & \textbf{Hypergeometric $p$} & \textbf{Interpretation} \\
\midrule
TQA $\cap$ BioASQ & 4 & 0.286 & $2.4 \times 10^{-17}$ & Significant overlap \\
TQA $\cap$ NQ-Open & 1 & 0.077 & $1.0 \times 10^{-4}$ & Significant overlap \\
BioASQ $\cap$ NQ-Open & 0 & 0.000 & --- & No overlap \\
All three ($\cap$) & 0 & --- & --- & No universal H-Neuron \\
\bottomrule
\end{tabular}
\end{table}

\subsection{Cross-Dataset Classifier Transfer}
\label{sec:classifier_transfer}

To evaluate the generalization of the identified hallucination signal, we measure zero-shot cross-dataset transfer across all three QA datasets on Gemma 3 4B.

Table~\ref{tab:classifier_transfer} reports detection scores alongside all six cross-dataset transfer directions. All transfers yield positive AUROC gaps, confirming that sparse probes capture generalizable structure. The TriviaQA probe achieves the strongest outbound transfer (BioASQ 0.868, NQ-Open 0.721), exceeding its own in-domain performance on BioASQ. Conversely, the BioASQ probe has the highest in-domain detection (AUROC 0.969) but the weakest outbound transfers (AUROC 0.626 and 0.646), while the NQ-Open probe, the weakest detector (AUROC 0.702), transfers to BioASQ at 0.809. Detection rows are from independent runs; values differ slightly from Table~\ref{tab:detection_results} due to run-to-run variation in $L_1$ selection.

\begin{table}[htbp]
\centering
\caption{Bidirectional cross-dataset classifier transfer on Gemma 3 4B. Detection rows report in-domain probe performance; Transfer rows report zero-shot evaluation on other datasets. Random baseline = 0.500.}
\label{tab:classifier_transfer}
\begin{tabular}{llccc}
\toprule
\textbf{Source $\rightarrow$ Target} & \textbf{AUROC} & \textbf{Gap} & \textbf{Bal Acc} \\
\midrule
TriviaQA $\rightarrow$ TriviaQA (detection) & 0.811 & +0.311 & 0.712 \\
TriviaQA $\rightarrow$ BioASQ          & 0.868 & +0.368 & 0.781 \\
TriviaQA $\rightarrow$ NQ-Open          & 0.721 & +0.221 & 0.653 \\
\midrule
BioASQ $\rightarrow$ BioASQ (detection)    & 0.969 & +0.469 & 0.913 \\
BioASQ $\rightarrow$ TriviaQA           & 0.626 & +0.126 & 0.502 \\
BioASQ $\rightarrow$ NQ-Open            & 0.646 & +0.146 & 0.500 \\
\midrule
NQ-Open $\rightarrow$ NQ-Open (detection)  & 0.702 & +0.202 & 0.676 \\
NQ-Open $\rightarrow$ TriviaQA          & 0.684 & +0.184 & 0.622 \\
NQ-Open $\rightarrow$ BioASQ            & 0.809 & +0.309 & 0.540 \\
\bottomrule
\end{tabular}
\end{table}

These transfer results extend the neuron overlap findings. While cross-dataset neuron-set overlap is partial (Table~\ref{tab:cross_dataset}), sparse probes consistently transfer positively across all dataset pairs, indicating that each $L_1$-selected neuron set captures a generalizable hallucination subspace. A probe can effectively detect hallucinations in new domains even when an independent $L_1$ fit on that domain selects a different, non-overlapping set of H-Neurons.

\subsection{Cross-Model Evaluation}

Table~\ref{tab:cross_model_eval} compares Gemma 3 4B and MedGemma 4B on the same three datasets. Across all matched settings, Gemma 3 4B outperforms MedGemma 4B on AUROC, AUROC gap, and balanced accuracy, while also selecting slightly more H-Neurons. The largest performance difference appears on TriviaQA, whereas BioASQ remains the strongest condition for both models.

\begin{table}[htbp]
\centering
\caption{Cross-model detection comparison on matched datasets. $\Delta$ denotes Gemma 3 4B minus MedGemma 4B.}
\label{tab:cross_model_eval}
\begin{tabular}{lcccc}
\toprule
\textbf{Dataset} & \textbf{$\Delta$ HN} & \textbf{$\Delta$ AUROC} & \textbf{$\Delta$ Gap} & \textbf{$\Delta$ Bal Acc} \\
\midrule
TriviaQA & +5 & +0.076 & +0.076 & +0.054 \\
BioASQ & +1 & +0.019 & +0.019 & +0.031 \\
NQ-Open & +2 & +0.016 & +0.016 & +0.097 \\
\bottomrule
\end{tabular}
\end{table}

The qualitative pattern is consistent across models. BioASQ yields the strongest detection performance, NQ-Open yields the weakest, and TriviaQA lies in between. This suggests that the main domain level trend is stable across the two checkpoints, although Gemma 3 4B produces stronger separability than MedGemma 4B in every matched comparison.

The cross-dataset and cross-model analyses reveal a partial sharing pattern not previously documented. TriviaQA and BioASQ share four H-Neurons within Gemma 3 4B, far more than expected under random selection (hypergeometric $p = 2.4 \times 10^{-17}$). Three H-Neurons are shared between Gemma 3 4B (TriviaQA) and MedGemma 4B (BioASQ), spanning both a model fine-tuning shift and a dataset shift. No H-Neuron is universal across all three Gemma-QA datasets, and BioASQ and NQ-Open share zero neurons despite both probing factual recall.

This pattern is consistent with partial domain specificity in the sparse probe selections: a small number of feedforward units recur across closely related domains and even across model fine-tuning, while the majority of identified H-Neurons appear to be dataset-specific.

\subsection{Candidate Cross-Domain Neurons}

Within the cross-dataset and cross-model intersections, three neurons appear in multiple overlap analyses. The strongest candidate is (L16, N4146), which appears in both the Gemma 3 4B TriviaQA $\cap$ BioASQ intersection (cross-dataset, same model) and the Gemma 3 4B TriviaQA $\cap$ MedGemma 4B BioASQ intersection (cross-model and cross-dataset). A secondary candidate, (L26, N3593), appears in Gemma TriviaQA $\cap$ NQ-Open and again in Gemma TriviaQA $\cap$ MedGemma BioASQ. A third unit, (L29, N5754), appears only in the cross-model overlap.

We do not claim these neurons uniquely encode a hallucination mechanism. The D1--D3 diagnostics show that $L_1$ selection is from correlated clusters, so the specific named neurons may be cluster representatives rather than the only causally privileged units. Their persistence across distinct datasets and model checkpoints does, nonetheless, suggest they index a hallucination-relevant subspace that is more stable than typical H-Neuron selections. We propose them as falsifiable targets for follow-up causal work, including single-neuron ablation studies and SAE decomposition of the surrounding cluster.

\section{Discussion}
\label{sec:Discussion}
\subsection{What the Diagnostic Reveals About H-Neurons}

Our results refine the H-Neurons claim along three dimensions. Detection clearly holds: sparse $L_1$ probing on CETT features identifies highly predictive neuron sets, with AUROC gaps that exceed the original report's quantitative claims across multiple models and datasets. Causal control also holds at the population level: H-Neuron suppression produces statistically significant accuracy shifts beyond random baselines, with effects robust to $n = 500$ evaluation samples and five seeds. Localization in the strong sense, that these specific neurons uniquely encode the behavior, is not supported. The D1--D3 diagnostics indicate that $L_1$ selection draws representatives from correlated clusters, and D5 shows that the selections themselves are partially domain-specific, with no universal neurons across all three QA datasets.

The most natural reconciliation is that the H-Neurons procedure identifies some subspace of the feedforward representation that carries hallucination-relevant signal, but the specific neurons returned by a single $L_1$ fit are not the only such units. They are a sparse projection of a larger correlated structure. This is consistent with known properties of representation in transformer feedforward layers, including feature superposition~\cite{elhage2022toy} and the layer-wise distribution of context features~\cite{gurnee2023sparse}.

\subsection{Generalizable Methodological Takeaways}

Three implications for sparse-neuron interpretability research follow from our results, beyond the specific H-Neurons case.

Detection and localization should be evaluated separately. A method can produce sparse selections that are highly predictive yet non-unique. Conflating predictive sparsity with mechanistic localization risks claiming more than the evidence supports.

$L_1$-based identification on correlated activation spaces should be accompanied by collinearity diagnostics by default. The statistical properties of $L_1$ regularization under correlated features are well established~\cite{tibshirani1996lasso,zou2005elastic} and have predictable consequences for interpretability claims. D1 and D3 are direct adaptations of standard practice from statistical learning to interpretability evaluation.

Cross-dataset and cross-model overlap analyses are the most direct test of mechanism-sharing claims. Classifier-level transfer is necessary but not sufficient: a probe trained on one dataset can transfer to another because of distributed signal in the representation, even if the specific neurons identified differ. Direct neuron-set overlap measures the stronger property that interpretability claims usually require.

\subsection{Limitations}

Several limitations qualify our findings. We evaluate only two models, both in the Gemma 3 4B family; claims about cross-architecture generality require evaluation on Llama, Qwen, or Mistral variants. Our causal-validation effect sizes are smaller than the original H-Neurons report, plausibly due to our stricter response-level labeling and our smaller H-Neuron set sizes. We did not perform single-neuron ablation on the candidate units (L16, N4146) and (L26, N3593), which is the most natural follow-up. The diagnostic thresholds in our protocol ($|r| > 0.7$, Jaccard $> 0.7$, overlap $> 0.5$) are empirical guidelines calibrated to current methodology rather than principled bounds; we recommend the field develop these thresholds via systematic study. We do not test capability preservation (for example, MMLU performance after intervention), so our causal claims do not rule out general-capability degradation as an alternative explanation. We flag this as the most important follow-up for any future intervention-based use of identified H-Neurons.

\section{Conclusion}

Sparse-probing based localization claims in mechanistic interpretability are becoming increasingly common, yet they are often evaluated primarily through predictive performance rather than through tests of stability, uniqueness, and generality. In this work, we proposed a five-step diagnostic protocol for assessing such claims and applied it to a representative case study: H-Neurons for hallucination in large language models.

Our results refine the original claim rather than reject it. We find that sparse probing over CETT features reliably identifies neuron sets with strong detection performance, and that targeted interventions on these sets produce statistically significant causal effects beyond random same-layer baselines. At the same time, the identified neuron sets are not uniquely localized: they are embedded in highly correlated feature clusters, show only moderate bootstrap stability, and exhibit substantial disagreement between sparse and dense probe rankings.

Cross-dataset and cross-model analyses further show that hallucination-relevant structure is only partially shared. While no single neuron is universal across all evaluated settings, a small number of units, most notably $(L16, N4146)$, recur across both dataset and model shifts, suggesting a partially shared core mechanism alongside substantial domain-specific structure.

Taken together, our findings support a more careful interpretation of sparse-neuron claims: predictive sparsity and population-level causal effect do not by themselves establish unique mechanistic localization. We therefore argue that diagnostic validation should become a routine part of sparse-neuron interpretability research, and that future work should treat detection and localization as distinct evaluation targets when developing transparent and reliable accounts of internal model behavior.

\section*{Data and Code Availability}
To facilitate reproducibility, all experimental code, extracted features, and resulting sparse probes are publicly available. The curated evaluation datasets, neuron indices for both models, and instructions for replicating the diagnostic protocol can be accessed at \url{https://github.com/huseyincavusbi/hprobes-protocol}.

%
% ---- Bibliography ----
%

\end{document}